\documentclass[conference]{IEEEtran}
\IEEEoverridecommandlockouts

\usepackage[T1]{fontenc}
\usepackage{cite}
\usepackage{amsmath,amssymb,amsfonts}
\usepackage{algorithm}
\usepackage{algpseudocode}
\usepackage{graphicx}
\usepackage{subcaption}
\usepackage{textcomp}
\usepackage{xcolor}
\usepackage{booktabs}
\usepackage{multirow}
\usepackage{catchfilebetweentags}
\usepackage{tabularx}
\usepackage[flushleft]{threeparttable}

\def\BibTeX{{\rm B\kern-.05em{\sc i\kern-.025em b}\kern-.08em
    T\kern-.24em\lower.7ex\hbox{E}\kern-.125emX}}
\begin{document}

\title{FedPPA: Progressive Parameter Alignment for Personalized Federated Learning}

\author{
\IEEEauthorblockN{
    Maulidi Adi Prasetia\IEEEauthorrefmark{1},
    Muhamad Risqi U. Saputra\IEEEauthorrefmark{2}, and
    Guntur Dharma Putra\IEEEauthorrefmark{1}\IEEEauthorrefmark{3}
}
\IEEEauthorblockA{
    \IEEEauthorrefmark{1}Universitas Gadjah Mada, Indonesia
    \IEEEauthorrefmark{2}Monash University, Indonesia\\
    maulidiadiprasetia2000@mail.ugm.ac.id, risqi.saputra@monash.edu, gdputra@ugm.ac.id
}
\thanks{\IEEEauthorrefmark{3}Corresponding author.}
}


\maketitle

\begin{abstract}
Federated Learning (FL) is designed as a decentralized, privacy-preserving machine learning paradigm that enables multiple clients to collaboratively train a model without sharing their data. In real-world scenarios, however, clients often have heterogeneous computational resources and hold non-independent and identically distributed data (non-IID), which poses significant challenges during training. Personalized Federated Learning (PFL) has emerged to address these issues by customizing models for each client based on their unique data distribution. Despite its potential, existing PFL approaches typically overlook the coexistence of model and data heterogeneity arising from clients with diverse computational capabilities. To overcome this limitation, we propose a novel method, called Progressive Parameter Alignment (FedPPA), which progressively aligns the weights of common layers across clients with the global model's weights. Our approach not only mitigates inconsistencies between global and local models during client updates, but also preserves client's local knowledge, thereby enhancing personalization robustness in non-IID settings. To further enhance the global model performance while retaining strong personalization, we also integrate entropy-based weighted averaging into the FedPPA framework. Experiments on three image classification datasets, including MNIST, FMNIST, and CIFAR-10, demonstrate that FedPPA consistently outperforms existing FL algorithms, achieving superior performance in personalized adaptation.
\end{abstract}


\section{Introduction}
Federated Learning (FL) is a distributed machine learning framework in which multiple clients collaboratively train a shared model without directly sharing their data, thereby enhancing overall model performance while preserving data privacy~\cite{maadyFederatedLearningIndonesia2022}. 
FL works through iterative training rounds between participating clients and a central aggregation server. In each round, clients train models locally using their private data and transmit the resulting model to the server. Subsequently, the server aggregates these models to generate a global model, which is then redistributed to the clients for further local training in the next round~\cite{liuKeepYourData2021, hsuFederatedLearningUsing2024}. However, in real-world scenarios, participating clients often have incomplete or sparse data, i.e., non-independent and identically distributed data (non-IID), which would degrade the overall model performance.\par

Personalized Federated Learning (PFL) addresses these data heterogeneity issues in the clients by grouping the models into a shared base layer and a personalized head layer~\cite{arivazhaganFederatedLearningPersonalization2019}. The base layer, shared by the server and all clients, encodes the common deep learning layer of the entire clients, while the head layer is locally customized and is fine-tuned to capture the unique characteristics of the client's data. These personalization layers enable each client's model to more accurately align with local data while still gaining advantages from the collective knowledge shared across the networks.
While PFL yields strong performance, these schemes typically demand all clients to have identical architecture. However, in practice, clients often possess heterogeneous computational resources, making it impractical to enforce a uniform model architecture across all clients, leading to each client having distinct architectures. Consequently, it becomes challenging to implement this personalization strategy that demands consistent model structures.\par

Recent studies~\cite{wangFlexiFedPersonalizedFederated2023, wangFedADPUnifiedModel2025} proposed PFL methods to address the issue of heterogeneous model architectures between clients. FlexiFed identifies common components within clients’ local models and facilitates joint training on these layers to effectively fuse knowledge while maintaining architectural flexibility~\cite{wangFlexiFedPersonalizedFederated2023}. In addition, FedADP introduces a comprehensive model strategy to manage a variety of heterogeneous network architectures and ensures that devices with limited computational capacity can still significantly contribute to the global model's training~\cite{wangFedADPUnifiedModel2025}. However, these approaches often overlook the potential parameter misalignment between the server (global) model and the individual clients' model during local updates, potentially leading to inconsistencies between existing client's parameters and the updated parameters from the global model. To address this knowledge inconsistency, FedAS proposes a solution to align the global model’s parameter with those of each client during local training, mitigating inconsistencies and improving overall model performance~\cite{yangFedASBridgingInconsistency2024}. However, this solution heavily relies on the assumption that every client uses the same model architecture, which is quite unlikely given the significant variations in computational power commonly found in real-world scenarios.

In this paper, we propose FedPPA: a PFL approach to simultaneously address heterogeneity in client models and non-IID nature of client data, while also resolving the knowledge inconsistency between the server and the client models during local updates. FedPPA progressively aligns the weights of common layers across clients with the global model's weights, mitigating inconsistencies between global and local models during client updates. While our main target is to improve personalization effectiveness and address the knowledge discrepancy problem, our approach also preserves client's local knowledge, thereby enhancing personalization robustness in non-IID settings. In addition, we propose FedPPA+, in which we incorporate an entropy-based weighted averaging to FedPPA, retaining strong personalization with a better trade-off to global model performance. The entropy values are calculated from each client's variability in label distribution, where more labels exist in client's dataset equal to a higher contribution weight.

In summary, we propose the following contributions:
\begin{itemize}
    \item We develop a PFL approach, termed FedPPA, which progressively aligns client's common layers with the global model parameters during local updates, which is suitable for FL with heterogeneous model architectures and non-IID data distribution.
    \item We incorporate an entropy-based weighted averaging in the global aggregation mechanism to better adapt to non-IID data distribution.
    \item We perform a comprehensive experiment and analysis on publicly available data sets for image classification (i.e., MNIST, F-MNIST, and CIFAR-10) using heterogeneous model architectures (i.e., VGG-11, VGG-13, VGG-16, and VGG-19).
\end{itemize}

The remainder of this paper is structured as follows. We present the related work in Section~\ref{sec:related-work} and discuss our proposed solution in Section~\ref{sec:methodology}. The implementation of our solution with the corresponding evaluation is presented in Section~\ref{sec:results}, while the conclusion of the paper is presented in Section~\ref{sec:conclusion}.

\section{Related Work}
\label{sec:related-work}
FL was introduced as a modern distributed AI training paradigm designed to preserve user privacy by keeping data localized on user devices \cite{kairouzAdvancesOpenProblems2019}. Instead of sharing raw data with a central server, FL enables collaborative model training across multiple clients, ensuring high levels of data privacy and security. While FL has successfully introduced collaborative training across clients, achieving high model performance using this approach typically relies on the assumption that the datasets used for training are independent and identically distributed (IID). In real-world applications, IID assumption is rarely met, resulting in non-IID data distributions across clients, which can significantly degrade the performance of FL model. Previous studies have shown that non-IID data distribution can reduce FL model performance for up to 29\% \cite{efthymiadisAdvancedOptimizationTechniques2024}.\par

In order to address the performance degradation of standard FL model due to the non-IID data distribution among clients, PFL was introduced by Arivazhagan et al. \cite{arivazhaganFederatedLearningPersonalization2019} by incorporating personalization layers. These personalization layers allow each client's model to better fit local data while still benefiting from shared knowledge across the networks. Since then, various PFL methods have been developed to address the diverse challenges in distributed and personalized model training. For instance, Hierarchical PFL (HPFL) \cite{youHierarchicalPersonalizedFederated2023, wangEfficientHierarchicalPersonalized2024, gaoFederatedLearningService2024, chenSemiAsynchronousHierarchicalFederated2023, wangEfficientHeterogeneousMultiModal2024} was developed to enable hierarchical weight aggregation among groups of clients through edge servers, which act as an intermediate aggregation points. In this approach, clients with similar resources or data characteristics are clustered together, and they only allow to communicate with the server via the edge servers, avoiding direct communication of each client with the cloud server. This approach not only reduces communication costs and latency but also lessens the computational load on the central server, especially when training involves a large number of clients. However, clustering similar clients might lead to knowledge inconsistency across clusters due to limited information sharing, resulting in knowledge isolation. Additionally, the HPFL approach is best suited for edge computing architectures and may be challenging to be implemented in more general or less structured environments.\par

Conventional PFL personalizes each client’s model by dividing it into two parts: a shared base layer and a personalized head layer. While this approach has demonstrated good performance, it relies on the assumption that all clients use the same model architecture. In practice, however, clients often have varying computational resources, making it difficult to enforce a uniform model structure. To address this limitation, recent researches, such as~\cite{wangFlexiFedPersonalizedFederated2023, wangFedADPUnifiedModel2025, liuCollaborativeNeuralArchitecture2025, zhouPersonalizedFederatedLearning2024, yanPeachesPersonalizedFederated2024, xuCooperativeMultiModelTraining2025}, have begun exploring PFL methods that support heterogeneous model architectures across clients. One notable example is FlexiFed, proposed by \cite{wangFlexiFedPersonalizedFederated2023}. FlexiFed allows each client to use different model architectures during PFL training by introducing three model aggregation strategies. One such strategy, called Max-Common, clusters a subset of model architectures based on their common layers and aggregates weights only from these common layers. This approach facilitates effective federated training across clients with heterogeneous models, ensuring both flexibility and consistent performance. Nevertheless, this approach might overlook the possibility of parameter misalignment between the global model and each client’s local model during local updates, which can lead to inconsistencies in shared knowledge. Previous work, such as FedAS \cite{yangFedASBridgingInconsistency2024}, has attempted to resolve this problem by aligning the global model’s weights with those of the client during local training. However, this solution is limited to scenarios where clients share a uniform model architecture. In contrast, our work aims to tackle the challenges of PFL in the presence of both heterogeneous client models and non-IID data, while also resolving the knowledge inconsistency between the server and the client models during local updates.\par 

\begin{figure*}
    \centering
    \includegraphics[width=0.75\textwidth]{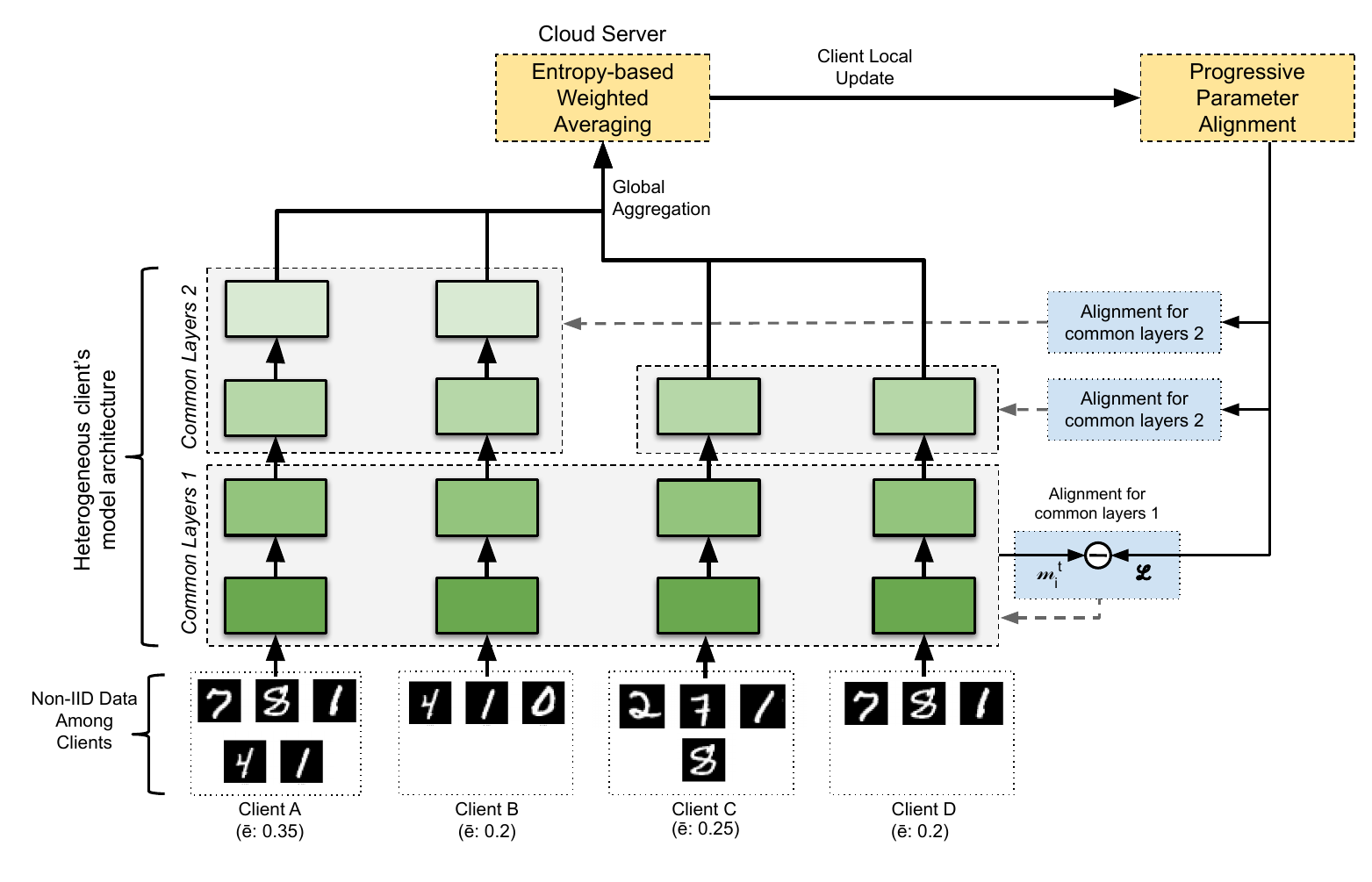}
    \caption{Overview of the proposed framework, where common layers are extracted from each client's heterogeneous models. Each client is assigned to an entropy-based contribution weight during global aggregation. The updated model is then returned to the clients, after which parameter alignment is progressively carried out locally based on each client's distinct dataset.
    }
    \label{fig:model_architecture}
\end{figure*}

\section{Methodology}
\label{sec:methodology}
Figure \ref{fig:model_architecture} shows an overview of our proposed framework that highlights our two key contributions (depicted as yellow box), namely 1) Progressive Parameter Alignment (PPA) and 2) entropy-based weighted averaging. Compared to the standard PFL methods, which straightforwardly override the client's base model with the global (server) model's weights during client update, our method progressively aligns common (and clustered) layers in the clients with the global model, addressing the knowledge-inconsistency problem between global and clients' models more effectively. By re-aligning the aggregated knowledge from the central server with each client’s local data before overwriting the client’s existing local knowledge, our method ensures that clients benefit from the latest server-aggregated information without sacrificing the personalized knowledge manifested in the local models. Note that this alignment occurs iteratively from the first common layer until the end, which matches and synchronizes the most similar layers as much as possible.

To better handle the non-IID data distribution, we employ an entropy‐based weighting to determine the contribution weight of each client based on the variability of its local data. Incorporating these weights yields more robust performance in highly skewed non-IID environments compared to the conventional PFL. The entropy-based weights ensure that clients with richer, more diverse datasets play a significant role in the knowledge aggregation process. In summary, the steps performed in each training round are described as follows:

\begin{itemize}
    \item \textbf{Calculation of client's weight}: In this step, we calculate client's weight contribution via the entropy-based method. Intuitively, we put higher weights to the clients with more diverse datasets.
    \item \textbf{Local training}: Each client trains their model locally using their personalized dataset. Note that the client's local dataset is divided into training and evaluation/testing dataset.
    \item \textbf{Clustering common layers}: All layers from each client are compared to identify a list of common layers. Both the architecture and the order of layers must match, e.g., clients with the same layer types but arranged differently are not considered to share common layers. Only clients containing the identified common layers would contribute to the aggregation process in the server.
    \item \textbf{Global aggregation}: In the server, the client's model parameters is aggregated based on the client's layer similarity. Note that the aggregation would leverage an entropy-based method, applied to the corresponding common layers.
    \item \textbf{Progressive parameter alignment}: Before updating each client's common layer parameters, parameter alignment process is performed. Using the list of common layers identified in the previous step, the previous client’s model features are extracted for each common layer. These output common features is used to align the client's model parameters by minimizing its differences with the global model features.
    \item \textbf{Model distribution}: In the final step, the aligned parameters of the common layers are distributed back to each client.
\end{itemize}

\subsection{Progressive Parameter Alignment --- FedPPA}
Conventional PFL methods typically split the client models into the base and the head (personalization) layers and overwrite their parameters directly during local updates. However, this approach could lead to inconsistent knowledge between local and global models, as the client already is trained with their personalized dataset. We introduce a progressive
parameter alignment approach, namely FedPPA, which aligns the aggregated global knowledge with client's personalized knowledge progressively for each common layer. In practice, FedPPA works by minimizing the differences between the global features and the features of the client model before performing local updates~\cite{yangFedASBridgingInconsistency2024}. The goal is to allow each client to share common knowledge for the same task without sacrificing personalized capabilities tailored to their unique dataset. 

We summarize the notations used in this paper in Table~\ref{tab:notation} and describe the complete workflow of FedPPA in Algorithm \ref{alg:fed_ppa_algorithm}. FedPPA begins by extracting the common layers of each client using the following function

\begin{table}[!htbp]
    \caption{Notations used in this paper and their definitions}
    \label{tab:notation}
    \begin{center}
        \begin{tabularx}{0.96\columnwidth}{cX|cX}
            \hline
            \textbf{Notation} & \textbf{Definition} & \textbf{Notation} & \textbf{Definition} \\
            \hline
            \hline
            $M$ & Number of clients & $C_l$ & Number of contributors for layer $l$ \\
            \hline
            $t$ & Number of rounds & $L_i$ & Total layers of client $i$ \\
            \hline
            $m^t_i$ & Client's model $i$ at round $t$ & $H^{(l)}_i$ & Output features of layer $l$ in client $i$ \\
            \hline
            $m^{t+1}_i$ & Updated client's model & $D_i$ & Dataset of client $i$ \\
            \hline
            $l$ & Layer of model $m$ & $e_i$ & Entropy value of client $i$ \\
            \hline
        \end{tabularx}
    \end{center}
\end{table}

\begin{equation}
    \label{eq:extract_max_common_layers}
    \begin{aligned}
        l^{t+1}_i, C_l = \mathrm{ExtractMaxCommonLayers}(m^t_i, m^t_{i+1}),
    \end{aligned}
\end{equation}
where $l^{t+1}_i$ is a set of common layers extracted on the client $i$ and $C_l$ represents the number of clients contributing to $l^{t+1}_i$. We adopt a layer-level extraction method from~\cite{wangFlexiFedPersonalizedFederated2023} into our $\mathrm{ExtractMaxCommonLayers}$ function. Subsequently, each model parameter from the extracted common layers is averaged using the following formula

\begin{equation}
    \label{eq:aggregated_local_model}
    \begin{aligned}
        m^{t+1}_i = \left\{\frac{1}{C_l} \sum_{l_i \in l^{t+1}_i} l_i\right\}^{L_i},
    \end{aligned}
\end{equation}
where $m^{t+1}_i$ is the updated parameter from client $i$ and $L_i$ is the number of layers in the model. This computation is performed iteratively for each layer in the client's local model. If
$l_i$ is selected as the common layers, it is incorporated into $l^{t+1}_i$, which means that the parameter (i.e., the "knowledge") from $l_i$ is added to the aggregated parameters. Before aligning the aggregated client's model parameters in each common layer with the global, we need to extract the current client's local features using the following formula

\begin{equation}
    \label{eq:feature_from_layer}
    \begin{aligned}
        H^{(l)}_i = \{h^{(l)}(x;m^t_i) | x \in D_i\} \in \mathcal{R}^{d_l \times N_i},
    \end{aligned}
\end{equation}
where $H^{(l)}_i$ is the feature extracted from layer $l$ in the local model $m^t_i$ using the personalized data set of the client $D_i$. This equation is also used to extract the features
from the updated and aggregated client model $m^{t+1}_i$ to ensure that the dimensions of the data are equal. Finally, each client's common layers are aligned with $m^t_i$ using the following equations

\begin{equation}
    \label{eq:align_averaged_layer}
    \begin{aligned}
        m^{t+1}_i \leftarrow m^{t+1}_i - \nabla\mathcal{L}(f(m^{t+1}_i, D_i) H^{(l)}_i)
    \end{aligned}
\end{equation}
and
\begin{equation}
    \label{eq:mse_loss}
    \begin{aligned}
        \mathcal{L}(f(m^{t+1}_i, D_i), H^{(l)}_i) = \|f(m^{t+1}_i, D_i) - H^{(l)}_i\|^2_2,
    \end{aligned}
\end{equation}
where $\mathcal{L}$ is the L2 loss function used to align the features between $m^{t+1}_i$ and $m^t_i$. The goal of this alignment is to make $m^{t+1}_i$ closer to $m^t_i$ by minimizing the differences of the features using Equation~\eqref{eq:mse_loss}.

\begin{algorithm}[b]
  \caption{FedPPA Algorithm}
  \label{alg:fed_ppa_algorithm}
  \begin{algorithmic}[1]
    \State $t$ = 1 \Comment{Round counter}
    \State $\mathcal{W}$ = $\varnothing$ \Comment{Store aligned models}
    \State

    \For{$i$ = 1,..,$M$}:
        \State $l^{t+1}_i, C_l$ = $\mathrm{ExtractMaxCommonLayers}$($m^t_i$, $m^t_{i+1}$) cf. \eqref{eq:extract_max_common_layers}
        \State Calculate $m^{t+1}_i$ cf. \eqref{eq:aggregated_local_model} \Comment{Aggregate model}
        \State Update $m^{t+1}_i$, cf. \eqref{eq:align_averaged_layer} \Comment{Parameter alignment}
        \State Add aligned model $m^{t+1}_i$ to $\mathcal{W}$
    \EndFor
    \State

    \State\Return $\mathcal{W}$
  \end{algorithmic}
\end{algorithm}

\subsection{Entropy-Based Client Weighting}
Conventional FL methods typically assume equal contribution from all clients during global aggregation. However, under non-IID data distributions and heterogeneous model architectures, client contributions should be weighted differently rather than treated uniformly. To this end, we proposed to leverage Entropy-based weighting to treat each client differently based on their dataset variability. Entropy, which quantifies the uncertainty or randomness of information, provides a natural measure for weighting each client’s influence. 


The Shannon's entropy~\cite{shannonMathematicalTheoryCommunication1948} method is employed to quantify how random and uncertain a piece of information is. In the context of a non-IID data distribution, a higher entropy value indicates that a client’s overall data are highly uncertain and diverse. To ensure fair comparison across clients, we normalize each client's entropy-based weight so that the total sums to one. Based on this procedure, we propose the following method for computing client contribution weights using the Shannon entropy formula


\begin{equation}
    \label{eq:entropy}
    \begin{aligned}
        e_i = \frac{\mathcal{H}(\mathcal{D})_i}{\sum_{i=1}^{M}\mathcal{H}(\mathcal{D})_i},
    \end{aligned}
\end{equation}
where $\mathcal{H}(\mathcal{D})_i$ is the Shannon Entropy of each client $i$, which can be calculated from \cite{lingFedEntropyEfficientFederated2023a,yeShannonEntropyQuasiparticle2024,lutzOptimizingFederatedLearning2024b}. As can be seen, each client's entropy value is divided with the sum of all client's entropy values to obtain the normalized entropy for each client, which is denoted as $e_i$.\par

\begin{figure*}[t]
	\centering
	\begin{subfigure}{0.25\linewidth}
		\includegraphics[width=\linewidth]{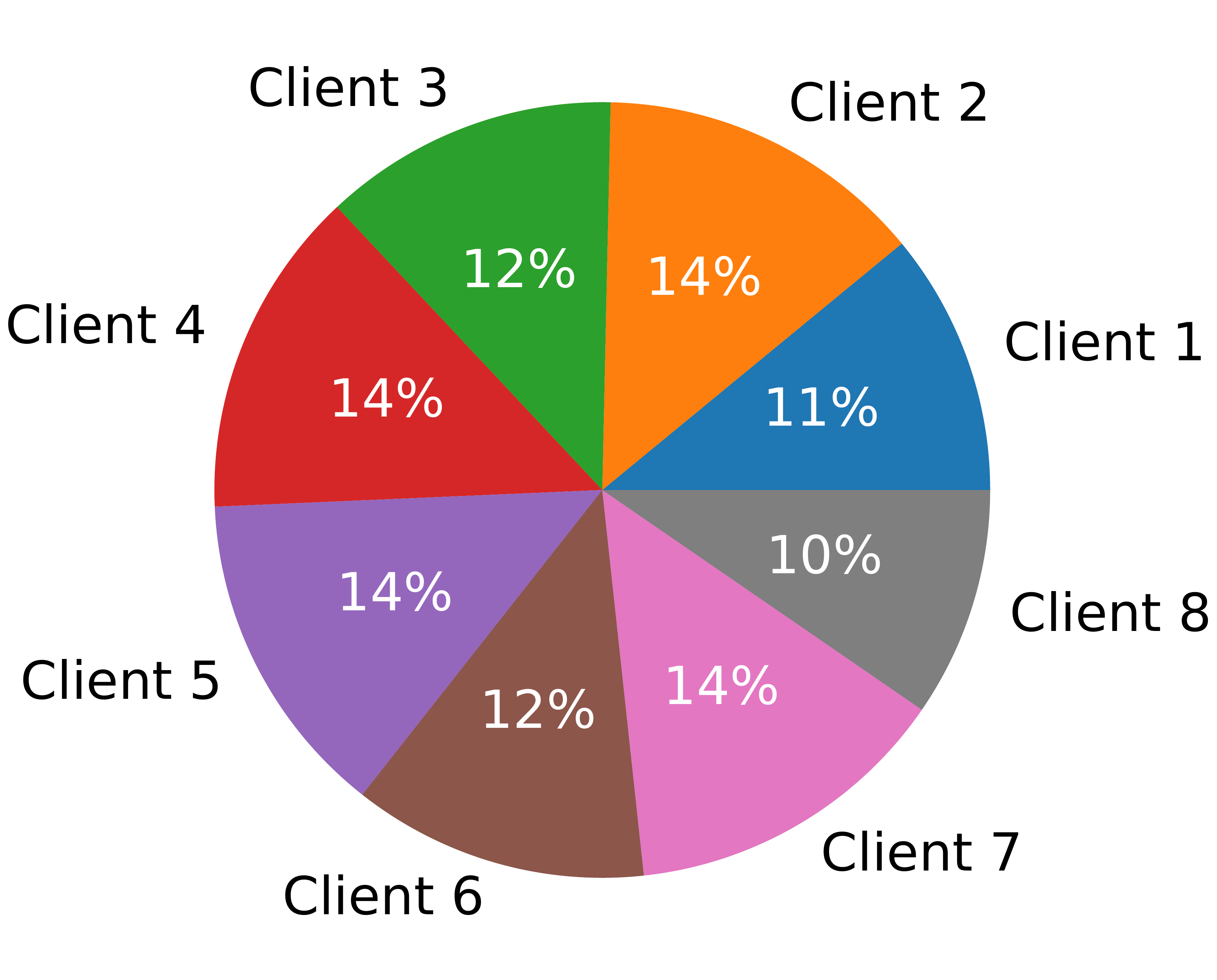}
        \caption{$\alpha=0.5$}
		\label{fig:subfigA}
	\end{subfigure}
	\begin{subfigure}{0.25\linewidth}
		\includegraphics[width=\linewidth]{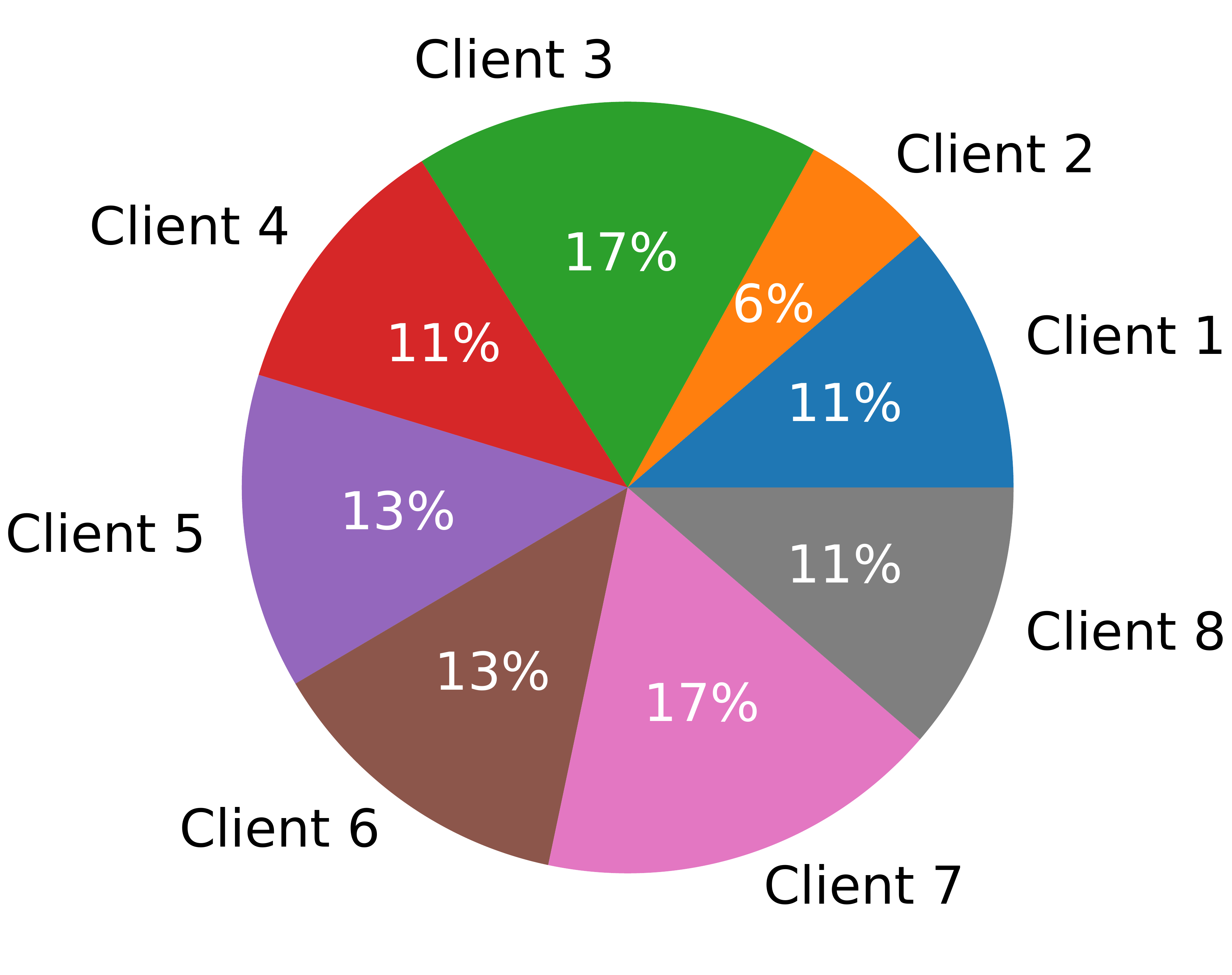}
        \caption{$\alpha=0.1$}
		\label{fig:subfigB}
	\end{subfigure}
	\begin{subfigure}{0.25\linewidth}
	    \includegraphics[width=\linewidth]{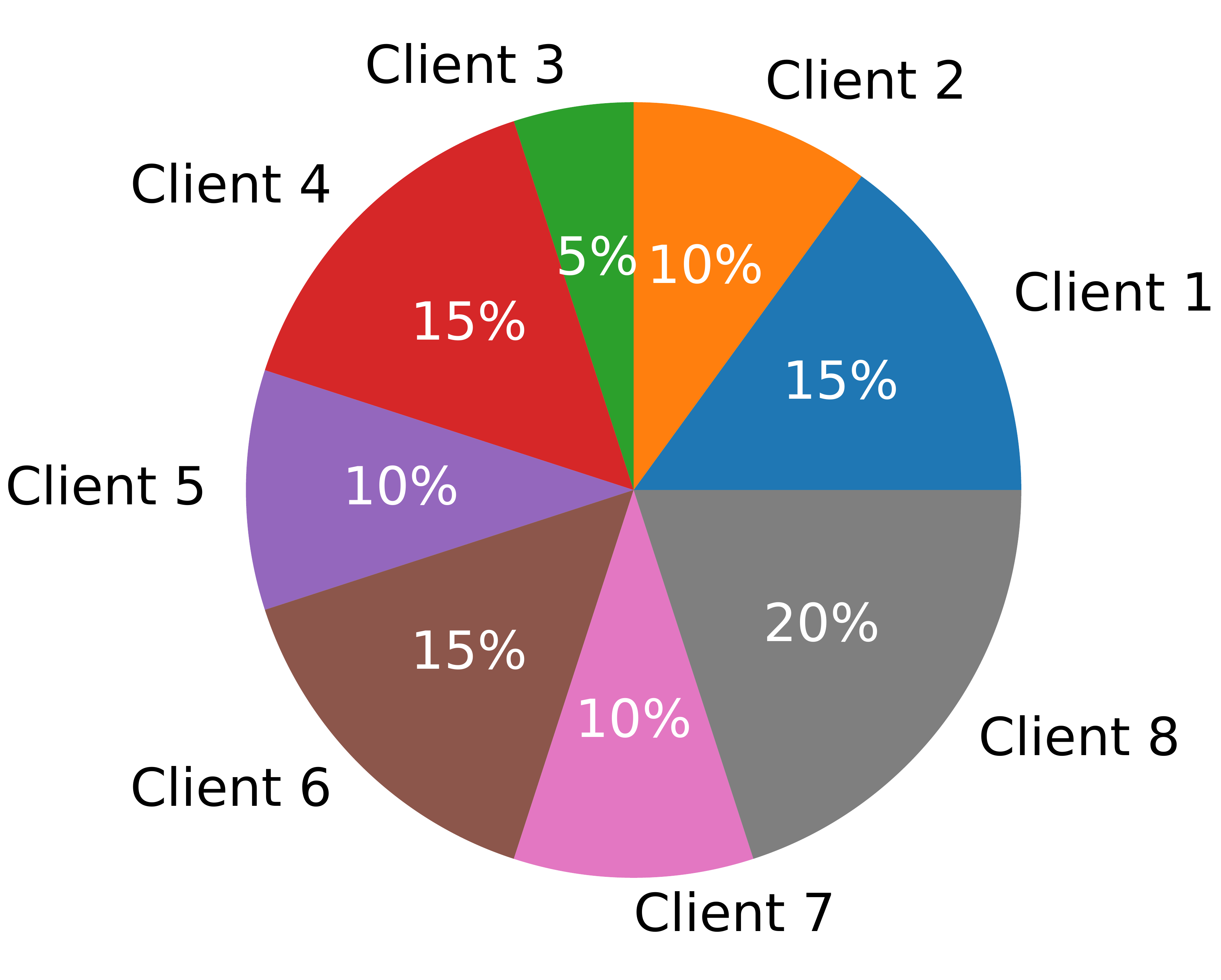}
        \caption{$\alpha=0.01$}
	    \label{fig:subfigC}
    \end{subfigure}

    \caption{Distribution of weighted contributions from each client, calculated using Shanon Entropy, applied to different non-IID data with different $\alpha$ values.}
	\label{fig:client_contribution}
\end{figure*}

Figure \ref{fig:client_contribution} shows the distribution of the contributions based on weighted entropy from each client, applied to different non-IID data with different Dirichlet values of $\alpha$ (see Section 4.2 for more information on how the non-IID data are generated). It can be seen that each client exhibits a distinct contribution weight that reflects the variability of its class labels. When a large value of $\alpha$ is used, the data distribution approaches IID, resulting in relatively uniform client contribution. On the other hand, when $\alpha$ decreases and the data become more non-IID, the variation in contribution weights across clients becomes more pronounced. This demonstrates that the entropy-based method effectively captures the degree of uncertainty and randomness in the client’s data.

\subsection{FedPPA with Entropy-Guided Weighted Averaging}
We propose FedPPA+, an extension of FedPPA that incorporates the client's entropy value as the weighting factor during the global knowledge aggregation. Each client would have
different aggregation weights based on their data distribution, as mentioned in the previous section. In general, FedPPA+ follows the same steps as FedPPA, which can be found in Algorithm \ref{alg:fed_ppa_plus_algorithm}, except that the global aggregation approach is now leveraging the entropy-based weighted averaging. The global aggregation process of FedPPA+ is formulated as follows:

\begin{equation}
    \label{eq:aggregated_local_model_w_entropy}
    \begin{aligned}
        m^{t+1}_i = \left\{e_i \times \left(\frac{1}{C_l} \sum_{l_i \in l^{t+1}_i} l_i\right)\right\}^{L_i}
    \end{aligned}
\end{equation}
where $e_i$ is the entropy value of client $i$. The only difference with FedPPA is that the averaged knowledge is not directly used as the new global model. Instead, it must first be weighted by $e_i$.

\begin{algorithm}[t]
  \caption{FedPPA+ Algorithm}
  \label{alg:fed_ppa_plus_algorithm}
  \begin{algorithmic}[1]
    \State $t$ = 1 \Comment{Round counter}
    \State $\mathcal{W}$ = $\varnothing$ \Comment{Store aligned models}
    \State

    \State Calculate $e$ for all clients using cf. \eqref{eq:entropy}
    \For{$i$ = 1,..,$M$}:
        \State $l^{t+1}_i, C_l$ = $\mathrm{ExtractMaxCommonLayers}$($m^t_i$, $m^t_{i+1}$) cf. \eqref{eq:extract_max_common_layers}
        \State Calculate $m^{t+1}_i$ with $e_i$ cf. \ref{eq:aggregated_local_model_w_entropy} \Comment{Aggregate model}
        \State Update $m^{t+1}_i$ cf. \ref{eq:align_averaged_layer} \Comment{Parameter Alignment}
        \State Add aligned model $m^{t+1}_i$ to $\mathcal{W}$
    \EndFor
    \State

    \State\Return $\mathcal{W}$
  \end{algorithmic}
\end{algorithm}

\section{Experimental Evaluation}
\label{sec:results}
\subsection{Dataset, Baselines, and Metrics}

\begin{figure*}[t]
  \centering
  \begin{subfigure}{0.32\textwidth}
    \centering
    \includegraphics[width=\textwidth]{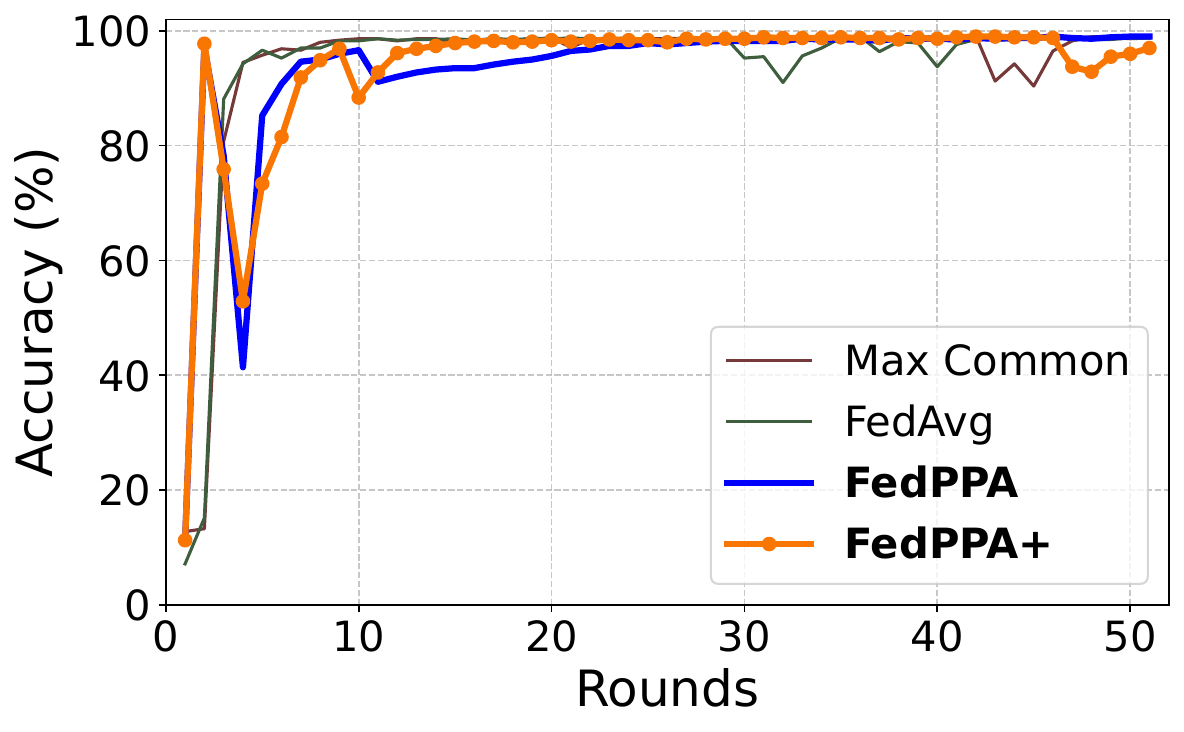}
    \caption{MNIST $\alpha=0.5$}
    \label{fig:mnist_0.5}
  \end{subfigure}
  \begin{subfigure}{0.32\textwidth}
    \centering
    \includegraphics[width=\textwidth]{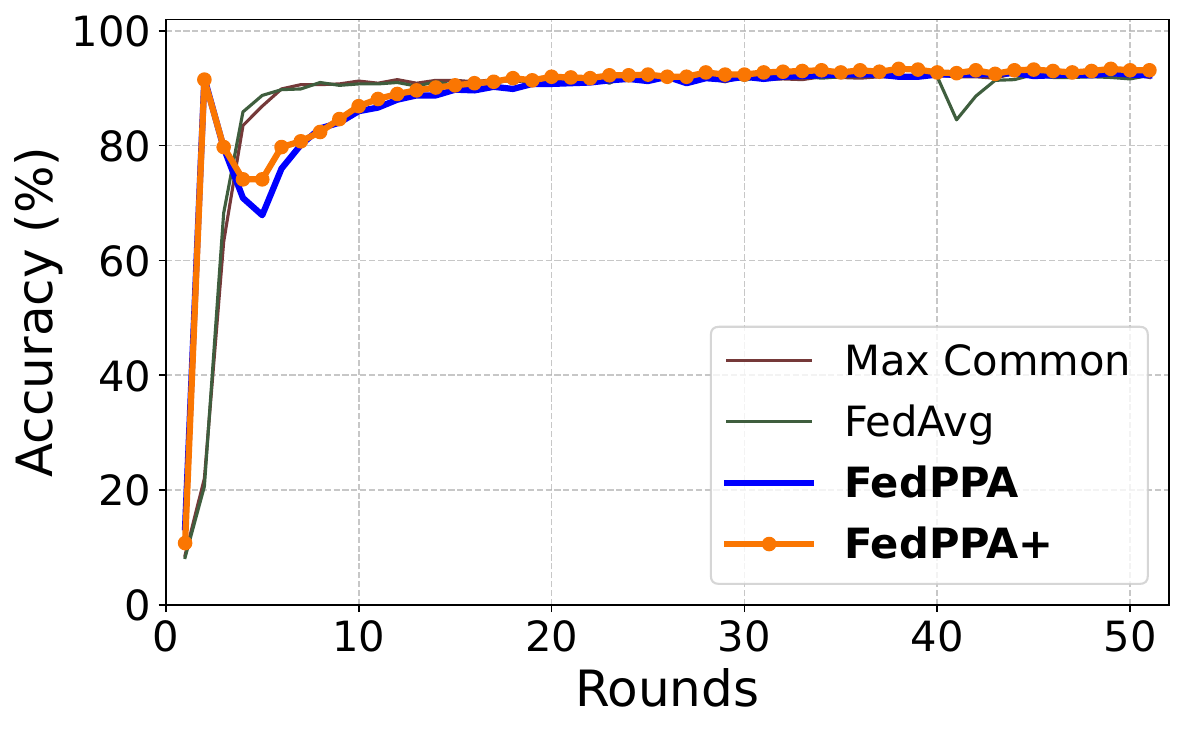}
    \caption{F-MNIST $\alpha=0.5$}
    \label{fig:f-mnist_0.5}
  \end{subfigure}
  \begin{subfigure}{0.32\textwidth}
    \centering
    \includegraphics[width=\textwidth]{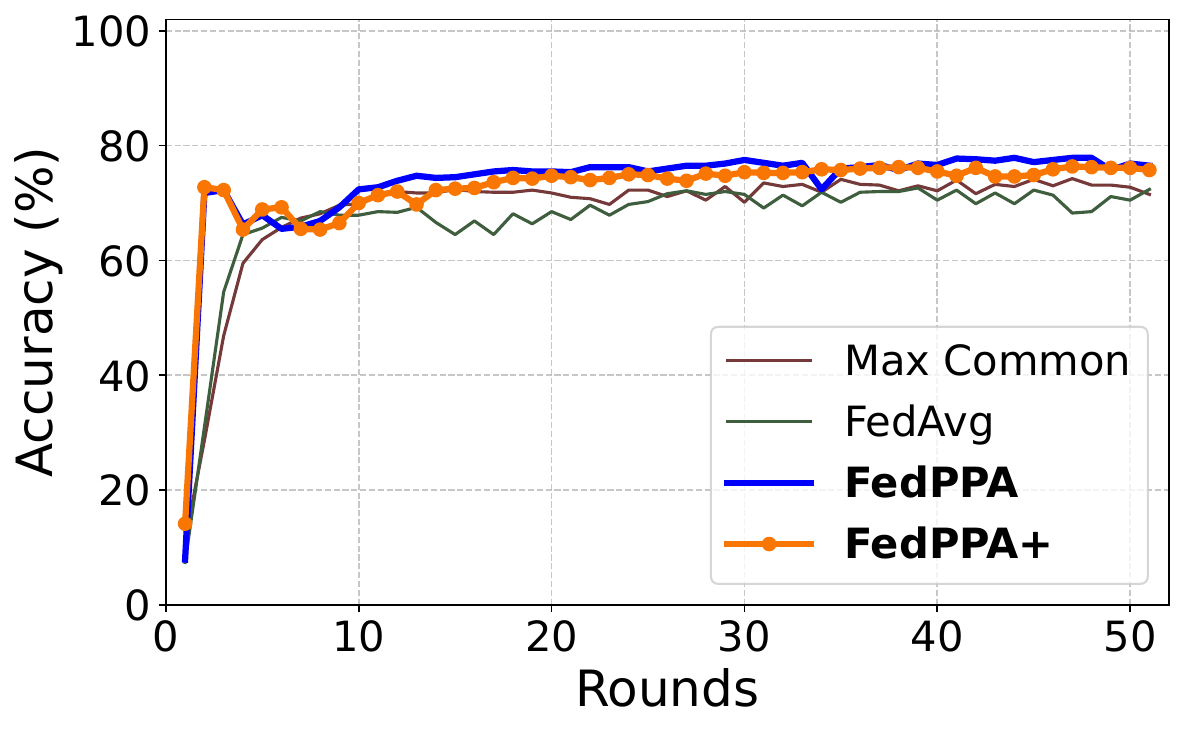}
    \caption{CIFAR-10 $\alpha=0.5$}
    \label{fig:cifar10_0.5}
  \end{subfigure}

  \caption{Model accuracy and convergence of our methods in Scenario 1 ($\alpha=0.5$) with two baseline models~\cite{wangFlexiFedPersonalizedFederated2023,mcmahanCommunicationEfficientLearningDeep2023}. While all methods can achieve good results, FedPPA and FedPPA+ surpass the baseline performances after 10 training iterations.}
  \label{fig:result_scenario_1}
\end{figure*}

\begin{figure*}[t]
  \centering
  \begin{subfigure}{0.32\textwidth}
    \centering
    \includegraphics[width=\textwidth]{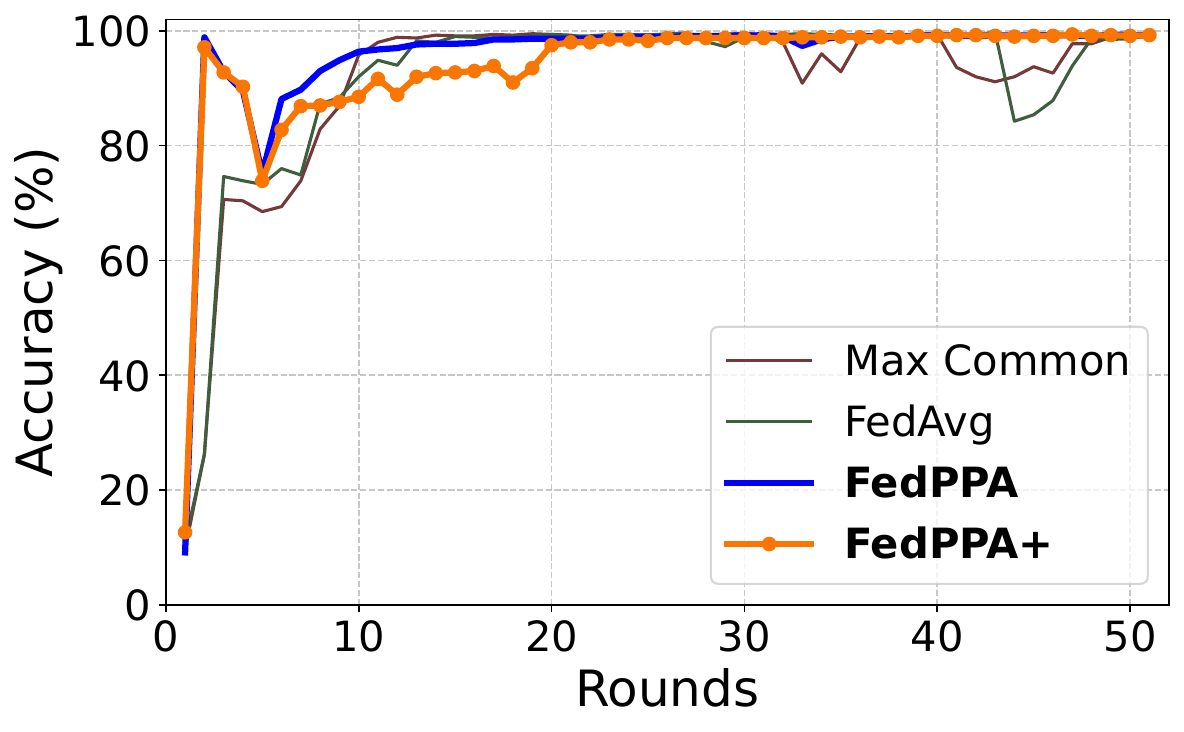}
    \caption{MNIST $\alpha=0.1$}
    \label{fig:mnist_0.1}
  \end{subfigure}
  \begin{subfigure}{0.32\textwidth}
    \centering
    \includegraphics[width=\textwidth]{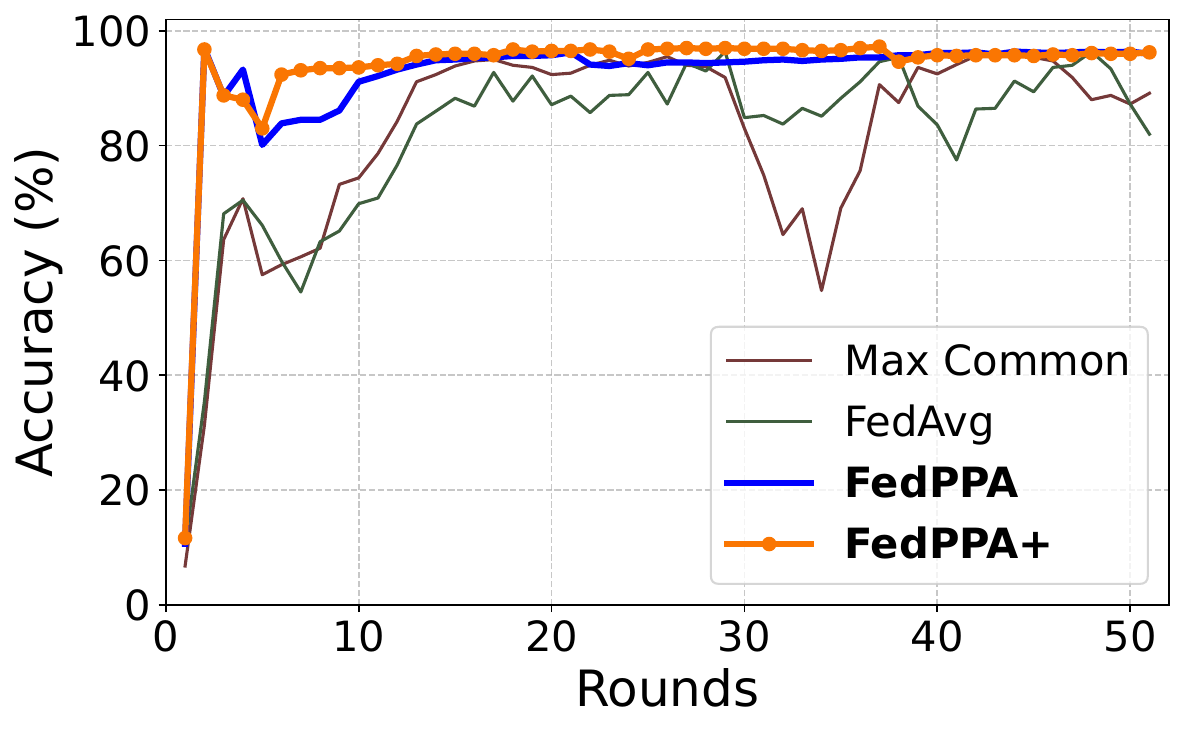}
    \caption{F-MNIST $\alpha=0.1$}
    \label{fig:f-mnist_0.1}
  \end{subfigure}
  \begin{subfigure}{0.32\textwidth}
    \centering
    \includegraphics[width=\textwidth]{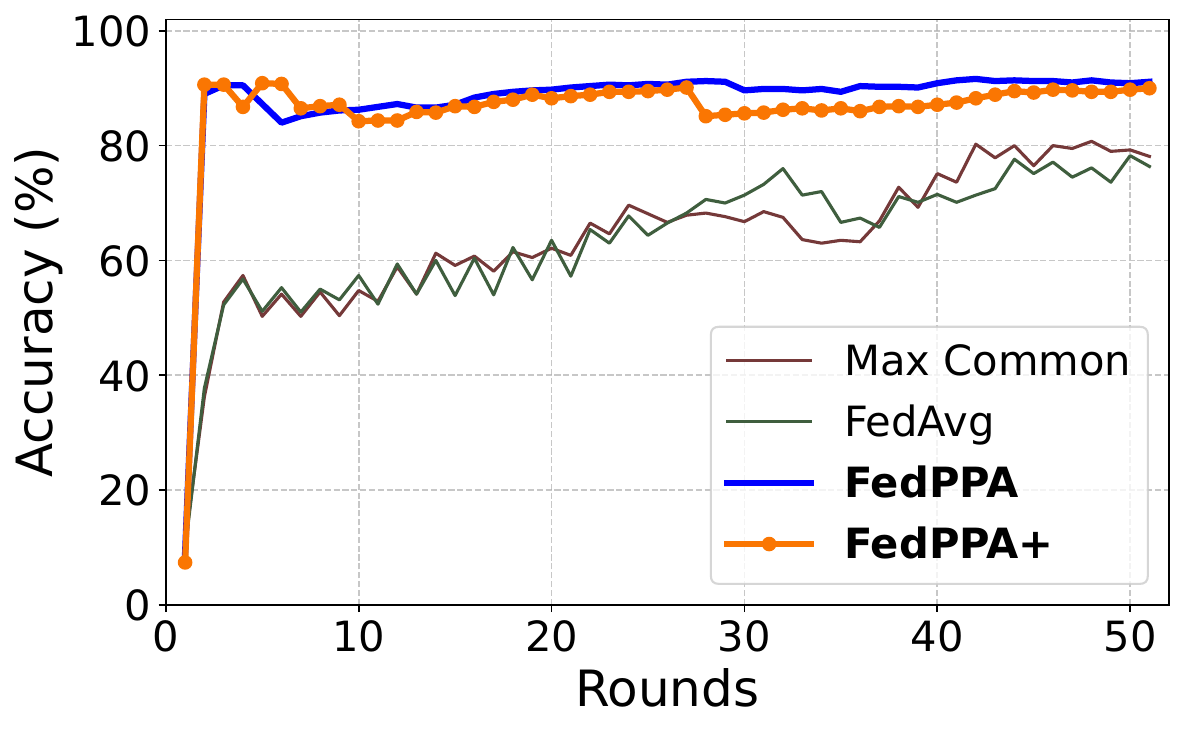}
    \caption{CIFAR-10 $\alpha=0.1$}
    \label{fig:cifar10_0.1}
  \end{subfigure}
  \caption{Model accuracy and convergence in Scenario 2 ($\alpha=0.1$). FedPPA and FedPPA+ exhibit stable performance relative to the baselines. As smaller $\alpha$ means more non-IIDness, our methods demonstrated superior performance in personalization.}
  \label{fig:result_scenario_2}
\end{figure*}

\begin{figure*}[t]
  \centering
  \begin{subfigure}{0.32\textwidth}
    \centering
    \includegraphics[width=\textwidth]{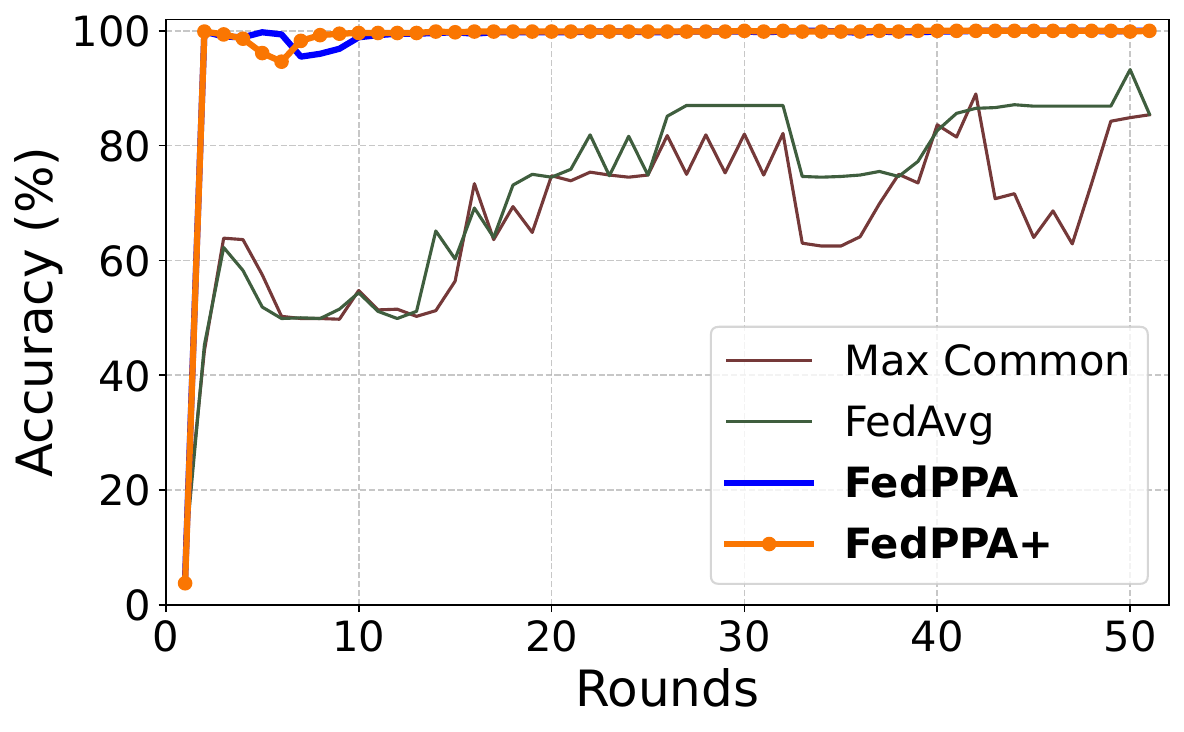}
    \caption{MNIST $\alpha=0.01$}
    \label{fig:mnist_0.01}
  \end{subfigure}
  \begin{subfigure}{0.32\textwidth}
    \centering
    \includegraphics[width=\textwidth]{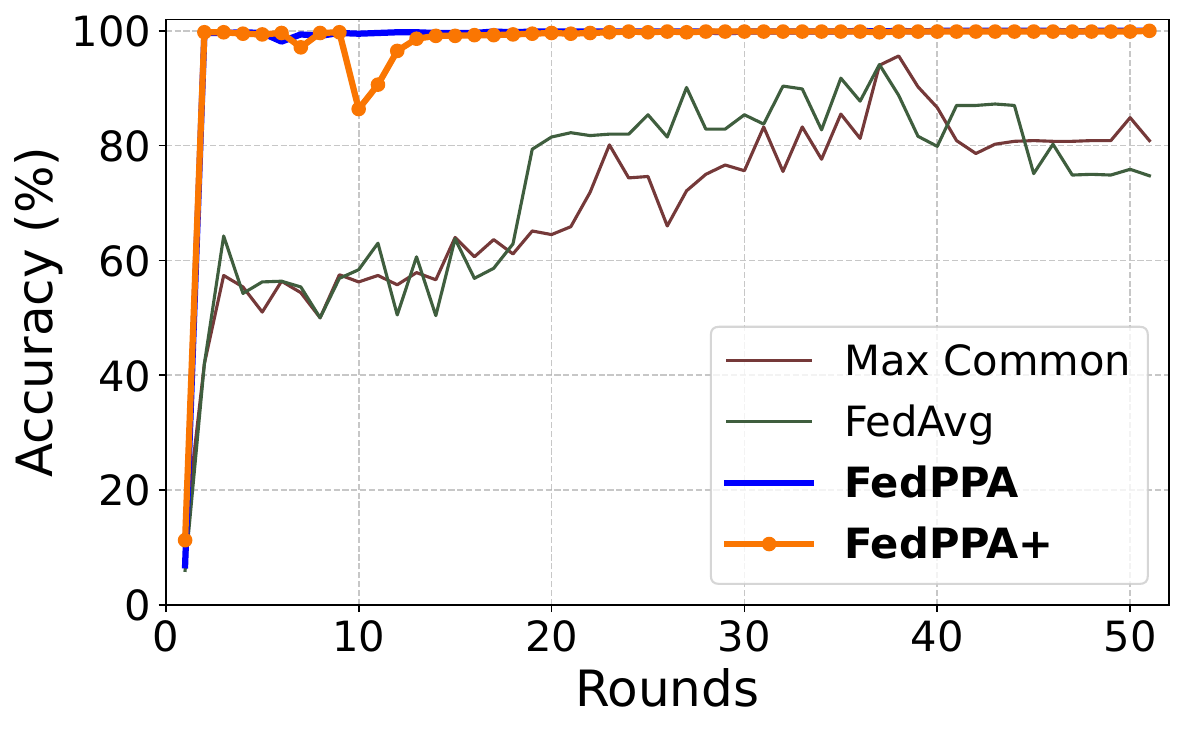}
    \caption{F-MNIST $\alpha=0.01$}
    \label{fig:f-mnist_0.01}
  \end{subfigure}
  \begin{subfigure}{0.32\textwidth}
    \centering
    \includegraphics[width=\textwidth]{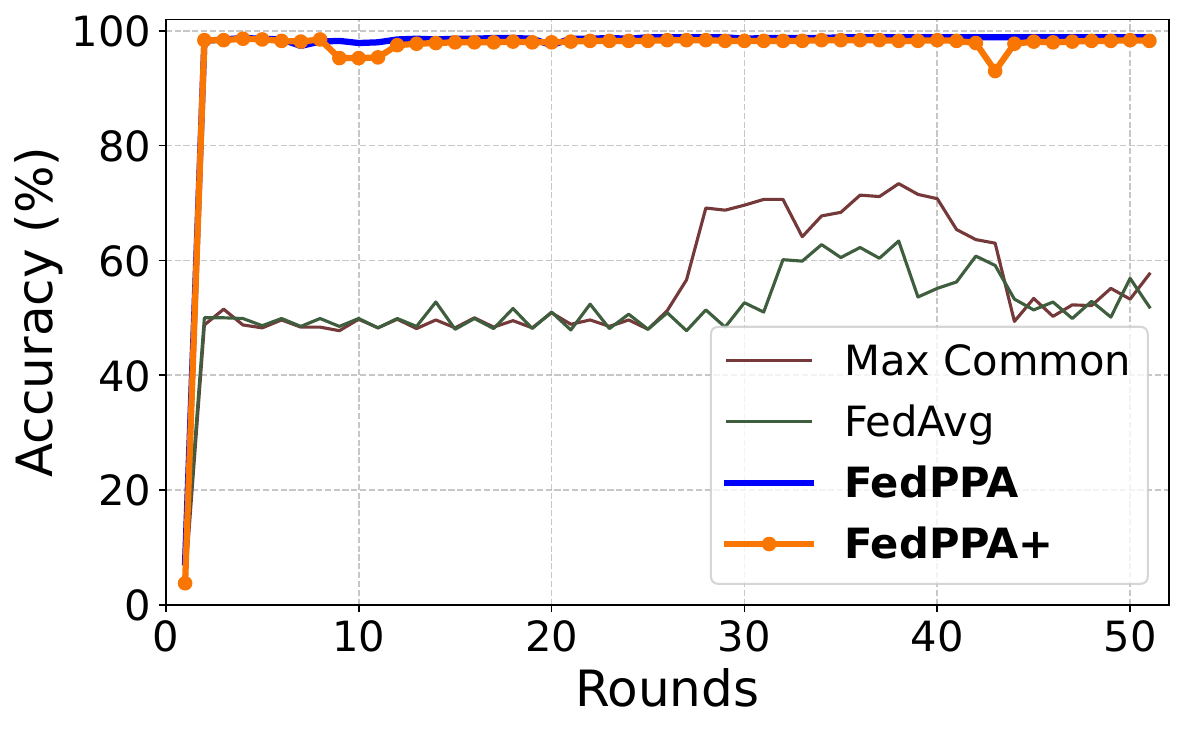}
    \caption{CIFAR-10 $\alpha=0.01$}
    \label{fig:cifar10_0.01}
  \end{subfigure}

  \caption{Model accuracy and convergence in Scenario 3 ($\alpha=0.01$). As $\alpha$ further decreases, the disparity between our methods and baseline models widens, highlighting the advantages of our proposed methods in handling personalization.}
  \label{fig:result_scenario_3}
\end{figure*}

We performed our experiments using publicly available dataset for image classification: 1) MNIST \cite{lecunMNISTDatabaseHandwritten}, a widely used dataset of 70K grayscale handwritten digit images (0–9); 2) Fashion-MNIST (F-MNIST) \cite{xiao2017/online}, a dataset of 70K grayscale images representing 10 categories of clothing items, serving as a more challenging version of MNIST; and 3) CIFAR-10 \cite{cifar-10}, which contains 60K 32×32 color images across 10 object categories and is considered as the most challenging dataset compared to MNIST and F-MNIST. As baselines, we compared our approach with Max-Common \cite{wangFlexiFedPersonalizedFederated2023} and FedAvg \cite{mcmahanCommunicationEfficientLearningDeep2023}, which represent the FL and PFL approaches, respectively. Finally, the classification accuracy was used as the primary evaluation metric to assess and compare the performance of each method. For a comprehensive evaluation, we measured both personalization performance (the average accuracy across individual clients) and global performance (the accuracy of the global model evaluated on the full dataset on the cloud server).


\begin{table}[b]
  \centering
  \caption{Experimental Scenario Setup}
  \newcolumntype{Y}{>{\centering\arraybackslash}X}
  \label{tab:scenario_experiment}
  \begin{tabularx}{\columnwidth}{c c *{3}{Y}}
    \hline
    \textbf{Scenario} & \textbf{Dirichlet $\alpha$} & \textbf{Dataset} & \textbf{Clients} & \textbf{Rounds} \\
    \hline
    \hline
    \multirow{3}{*}{1}  & \multirow{3}{*}{0.5}  & MNIST & \multirow{9}{*}{8}  & \multirow{9}{*}{51} \\ 
    &                     & F-MNIST  & &\\
    &                     & CIFAR-10 & &\\ 
    \cline{1-3}
    \multirow{3}{*}{2}  & \multirow{3}{*}{0.1}  & MNIST  \\
    &                     & F-MNIST  & & \\
    &                     & CIFAR-10 & & \\
    \cline{1-3}
    \multirow{3}{*}{3}  & \multirow{3}{*}{0.01} & MNIST  \\
    &                     & F-MNIST  & & \\
    &                     & CIFAR-10 & & \\
    \hline
  \end{tabularx}
\end{table}

\subsection{Experimental Setup}
\begin{figure*}[t]
	\centering
	\begin{subfigure}{0.3\linewidth}
		\includegraphics[width=\linewidth]{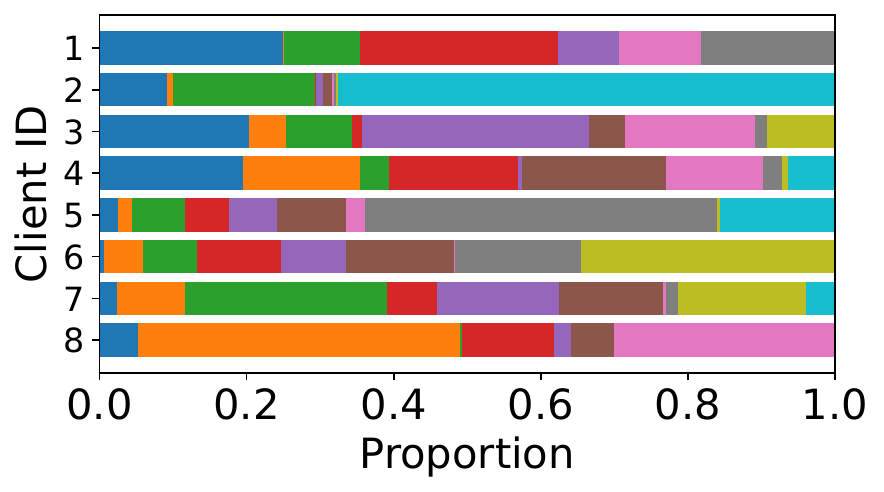}
        \caption{$\alpha = 0.5$}
		\label{fig:subfigA}
	\end{subfigure}
	\begin{subfigure}{0.3\linewidth}
		\includegraphics[width=\linewidth]{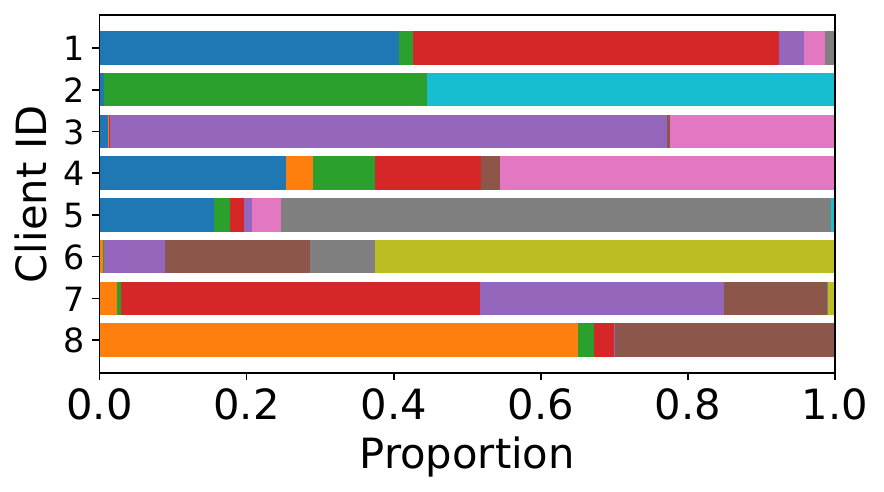}
        \caption{$\alpha = 0.1$}
		\label{fig:subfigB}
	\end{subfigure}
	\begin{subfigure}{0.3\linewidth}
	    \includegraphics[width=\linewidth]{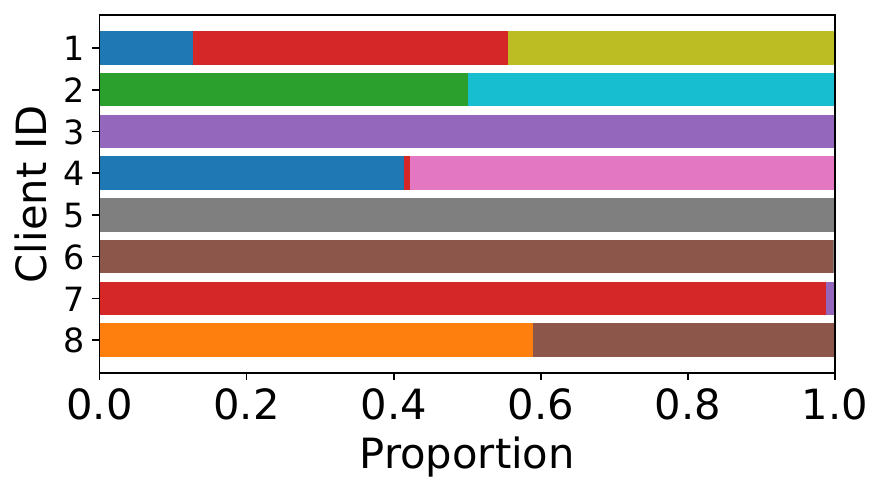}
        \caption{$\alpha = 0.01$}
	    \label{fig:subfigC}
    \end{subfigure}

    \caption{
    Distribution of labels among clients in the MNIST dataset using the Dirichlet Distribution approach.
    }
	\label{fig:label_contribution}
\end{figure*}

We conducted our experiments under conditions where clients had both non-IID data distributions and heterogeneous model architectures. Table \ref{tab:scenario_experiment} summarizes the experimental setup which shows the number of clients, the number of communication rounds, the dataset, and the $\alpha$ value.
To simulate a non-IID data distribution for each dataset, we leveraged the Dirichlet distribution-based method \cite{yangFedASBridgingInconsistency2024}, in which the parameter $\alpha$ is used to characterize the distribution, influencing the shape and concentration of probabilities between categories. Figure \ref{fig:label_contribution}
depicts the label distribution across clients with $\alpha \in \{ 0.5, 0.1, 0.01 \}$, where it represents label distributions from almost IID to highly non-IID. Note that each scenario is run for 10 epochs in each client's 
local training. \par

To simulate heterogeneous client model architectures, we utilized the VGG model for image classification with different depth, namely VGG-11, VGG-13, VGG-16, and VGG-19. Since we used a total of 8 clients and only 4 different model architectures were available, we assigned every 2 clients with the same model architecture. As one of the baselines, since FedAVG aggregates weights by averaging across client models, this implies that weight aggregation can only occur among clients sharing the same architecture. On the other hand, similar to our approach, Max-Common was executed as is, since the model has the ability to handle architecture heterogeneity.\par

We implemented our proposed framework using PyTorch library, where some of our implementations were inspired by the work of~\cite{yangFedASBridgingInconsistency2024}
and~\cite{wangFlexiFedPersonalizedFederated2023}. The experiment was conducted using an Nvidia RTX3060 GPU with 12GB of memory, a CUDA Version 12.3, Python v3.10.12, and PyTorch v2.2.0. The configuration includes 8 clients, each of whom locally trains their model for 10 epochs using the same hyperparameters. Stochastic Gradient Descent (SGD) was applied as an optimizer function with a learning rate of $0.01$, a momentum of $0.9$, and a learning rate decay of 5$e^{-4}$. The entire PFL training process was executed for $51$ rounds in total.\par

\subsection{Experimental Results}
Table \ref{tab:result_accuracy_personalized} shows the personalization performance of our approach, in terms of the average prediction accuracy, compared to the baselines in Scenario 1 ($\alpha=0.5$), 2 ($\alpha=0.1$), and 3 ($\alpha=0.01$), respectively. As we can see, in all datasets, our approaches, both FedPPA and FedPPA+, show superior personalization performance, except for MNIST with $\alpha=0.5$ although the differences with other approaches are relatively small (less than 0.5\%). As expected, Max-Common, representing a PFL approach, achieves overall better performance compared to FedAVG, which serves as a baseline for standard FL methods.

\begin{table}[b]
    \caption{Accuracy of Comparison in Personalized Evaluation}
    \label{tab:result_accuracy_personalized}
    \newcolumntype{Y}{>{\centering\arraybackslash}X}

    \centering
    \begin{threeparttable}
      \begin{tabularx}{\columnwidth}{c c *{4}{Y}}
        \hline
        \textbf{Dirichlet $\alpha$}
          & \textbf{Dataset}
          & \textbf{FedAvg} \,\,\,\,\,\, \cite{mcmahanCommunicationEfficientLearningDeep2023} & \textbf{Max-Common}\cite{wangFlexiFedPersonalizedFederated2023} 
          & \textbf{FedPPA} (Ours) & \textbf{FedPPA+} (Ours) \\
        \hline
        \hline
        \multirow{3}{*}{$\boldsymbol{0.5}$}
          & $\textbf{MNIST}$    & 99.75 & 99.62 & \textcolor{red}{99.37} & \textcolor{red}{99.37}   \\
          & $\textbf{F-MNIST}$  & 92.37 & 92.62 & \textcolor[cmyk]{0.87, 0, 0.94, 0.14}{92.62} & \textcolor[cmyk]{0.87, 0, 0.94, 0.14}{93.37}   \\
          & $\textbf{CIFAR-10}$  & 72.62 & 74.25 & \textcolor[cmyk]{0.87, 0, 0.94, 0.14}{77.87} & \textcolor[cmyk]{0.87, 0, 0.94, 0.14}{76.37}   \\ 
        \hline
        \multirow{3}{*}{$\boldsymbol{0.1}$}
          & $\textbf{MNIST}$    & 99.75 & 99.62 & \textcolor{red}{99.37} & \textcolor{red}{99.37}   \\
          & $\textbf{F-MNIST}$  & 96.37 & 96.24 & \textcolor[cmyk]{0.87, 0, 0.94, 0.14}{97.00} & \textcolor[cmyk]{0.87, 0, 0.94, 0.14}{97.25}   \\
          & $\textbf{CIFAR-10}$  & 78.25 & 80.75 & \textcolor[cmyk]{0.87, 0, 0.94, 0.14}{91.62} & \textcolor[cmyk]{0.87, 0, 0.94, 0.14}{90.87}   \\ 
        \hline
        \multirow{3}{*}{$\boldsymbol{0.01}$}
          & $\textbf{MNIST}$    & 93.25 & 89.00 & \textcolor[cmyk]{0.87, 0, 0.94, 0.14}{100.0} & \textcolor[cmyk]{0.87, 0, 0.94, 0.14}{100.0}   \\
          & $\textbf{F-MNIST}$  & 94.12 & 95.62 & \textcolor[cmyk]{0.87, 0, 0.94, 0.14}{100.0} & \textcolor[cmyk]{0.87, 0, 0.94, 0.14}{100.0}   \\
          & $\textbf{CIFAR-10}$  & 63.37 & 73.37 & \textcolor[cmyk]{0.87, 0, 0.94, 0.14}{98.87} & \textcolor[cmyk]{0.87, 0, 0.94, 0.14}{98.62}   \\ 
        \hline
      \end{tabularx}
      \begin{tablenotes}
        \item[1] \textcolor[cmyk]{0.87, 0, 0.94, 0.14}{Green color} indicates that our methods outperform the baselines.
        \item[2] \textcolor{red}{Red color} indicates that our method performance is below the baselines.
      \end{tablenotes}
    \end{threeparttable}
\end{table}

To gain a clearer understanding of each model’s performance over time during training, Figure~\ref{fig:result_scenario_1},~\ref{fig:result_scenario_2}, and~\ref{fig:result_scenario_3} illustrate the personalization performance of our approach across all scenarios, from the first until the last round. From these figures, our approaches generally converge faster than the baselines, achieving peak accuracy earlier in most scenarios. Notably, the performance gap between our model and the baselines widens as the value of $\alpha$ decreases (highly non-IID), reaching nearly 25\% differences in the most extreme case (See CIFAR-10 with $\alpha=0.01$ in Table \ref{tab:result_accuracy_personalized}). On the other hand, our model show only a slight performance improvement over the baselines at higher value of $\alpha$, where the data distribution is closer to IID, as seen in Figure \ref{fig:result_scenario_1}. This indicates that our model shows a strong robustness in personalization performance in an extreme non-IID scenario compared to the baselines. Finally, all compared models demonstrated relatively lower performance on the CIFAR-10 dataset, which is expected given that CIFAR-10 is considered more challenging than MNIST and F-MNIST.

The performance of our methods improved significantly as the level of non-IIDness increased in the dataset. As shown in Scenario 3 ($\alpha = 0.01$) of Table~\ref{tab:result_accuracy_personalized}, our approaches reach a prediction accuracy of up to 100\% for both MNIST and F-MNIST. In such extreme non-IID settings, some clients possess only one or two labels, as illustrated in Fig.~\ref{fig:label_contribution}(c). Under these circumstances, our highly personalized approach can achieve perfect predictions, given a minimal number of classes.

To highlight the performance differences between FedPPA and FedPPA+, Table~\ref{tab:result_global} presents the global performance evaluation between both methods across all tested datasets with $\alpha=0.5$. As shown, FedPPA+ outperforms FedPPA on two of the datasets, indicating its enhanced capability in handling non-IID data distributions. Although FedPPA+ achieves slightly lower accuracy on the MNIST dataset, the difference is minimal (less than 2\%). Nevertheless, further research is required to better balance personalized adaptation and global model performance, aiming to improve both personalization and global accuracy.


\begin{table}[b]
    \centering
    \caption{Global Performance of FedPPA and FedPPA+ with $\alpha=0.5$}
    \label{tab:result_global}
    \newcolumntype{Y}{>{\centering\arraybackslash}X}

    \begin{threeparttable}
    \begin{tabularx}{\columnwidth}{l l *{4}{Y}}
        \hline
        \multicolumn{2}{c}{Dataset} 
          & \multicolumn{2}{c}{FedPPA}
          & \multicolumn{2}{c}{FedPPA+} \\
        \cline{3-6}
        &  & Accuracy & Round & Accuracy & Round \\
        \hline
        \hline
          & \textbf{MNIST}     & 79.25 & 50 & \textcolor{red}{77.62} & 36     \\
          & \textbf{F-MNIST}   & 62.0  & 50 & \textcolor[cmyk]{0.87, 0, 0.94, 0.14}{63.0}  & 51     \\
          & \textbf{CIFAR-10}  & 38.87 & 49 & \textcolor[cmyk]{0.87, 0, 0.94, 0.14}{40.62} & 51     \\
        \hline
    \end{tabularx}
    \begin{tablenotes}
        \item[1] \textcolor[cmyk]{0.87, 0, 0.94, 0.14}{Green color} indicates that FedPPA+ performs better than FedPPA.
        \item[2] \textcolor{red}{Red colored} indicates that FedPPA+ performance is below FedPPA.
    \end{tablenotes}
    \end{threeparttable}
\end{table}

\textbf{Limitations.}
We acknowledge two primary limitations in the present study.
First, since FedPPA and FedPPA+ were evaluated only on VGG model variations, these evaluations may not capture unforeseen heterogeneity in other scenarios, which may yield different results, for instance, when the proposed methods are tested on transformer-based models.
Secondly, we conducted our evaluations on only eight FL nodes, which may require further exploration in scenarios involving hundreds of participating nodes to confirm the convergence of the model performance.
Further research may be directed towards validating these methods across more diverse model architectures and at a larger scale.

\section{Conclusion}
\label{sec:conclusion}
In this paper, we introduced a novel PFL method, termed FedPPA, designed to operate effectively in the presence of both heterogeneous client models and non-IID data, while also addressing the knowledge inconsistency between server and client models during local updates. FedPPA works by progressively aligning the weights of common layers between clients and the global model, thereby improving personalization performance under non-IID conditions. We further extended our approach with FedPPA+, which enhances model robustness by incorporating an entropy-based weighting mechanism during global model aggregation. Experimental results in three image classification datasets, namely MNIST, F-MNIST, and CIFAR-10, demonstrated that our methods consistently outperform baseline models in terms of personalization performance.
As FedPPA and FedPPA+ primarily aim to improve personalization effectiveness and address the issue of knowledge discrepancies, future work may be directed toward balancing personalization and global performance.

\section*{Acknowledgment}
We express our gratitude to the Indonesia Endowment Fund for Education (LPDP) under the Ministry of Finance of the Republic of Indonesia for providing the scholarship and supporting this research.

\bibliographystyle{ieeetr}
\bibliography{icsoc}

\end{document}